\documentclass[letterpaper, 10 pt, conference]{ieeeconf}  

\IEEEoverridecommandlockouts                              

\usepackage{multirow}
\usepackage{amsmath}
\usepackage{amssymb}
\usepackage{graphicx}
\usepackage[table]{xcolor}
\usepackage{tikz}
\usepackage{booktabs}
\let\labelindent\relax
\usepackage{enumitem}
\usepackage{xurl}
\makeatletter
\let\NAT@parse\undefined
\makeatother
\usepackage[
  colorlinks=true,
  linkcolor=blue,
  citecolor=blue,
  urlcolor=blue
]{hyperref}
\graphicspath{{./figures/}}

\definecolor{blueLighterRWTH}{cmyk}{0.45,0.14,0,0}
\definecolor{greenLightRWTH}{cmyk}{0.525,0,0.75,0}
\definecolor{orangeLightRWTH}{cmyk}{0,0.3,0.75,0}
\definecolor{redLightRWTH}{cmyk}{0.1125,0.75,0.75,0}
\definecolor{ikayellow}{cmyk}{0, 0.07, 1, 0}

\newif\ifblind
\blindfalse 

\title{\LARGE \bf \textbf{A Five-Layer MLOps Architecture for Connected Automated Driving}}

\usepackage{scalerel}
\usetikzlibrary{svg.path,backgrounds,arrows,shapes,fit,shadows,decorations,decorations.text,decorations.markings}

\definecolor{orcidlogocol}{HTML}{A6CE39}
\tikzset{
  orcidlogo/.pic={
    \fill[orcidlogocol] svg{M256,128c0,70.7-57.3,128-128,128C57.3,256,0,198.7,0,128C0,57.3,57.3,0,128,0C198.7,0,256,57.3,256,128z};
    \fill[white] svg{M86.3,186.2H70.9V79.1h15.4v48.4V186.2z}
                 svg{M108.9,79.1h41.6c39.6,0,57,28.3,57,53.6c0,27.5-21.5,53.6-56.8,53.6h-41.8V79.1z M124.3,172.4h24.5c34.9,0,42.9-26.5,42.9-39.7c0-21.5-13.7-39.7-43.7-39.7h-23.7V172.4z}
                 svg{M88.7,56.8c0,5.5-4.5,10.1-10.1,10.1c-5.6,0-10.1-4.6-10.1-10.1c0-5.6,4.5-10.1,10.1-10.1C84.2,46.7,88.7,51.3,88.7,56.8z};
  }
}
\newcommand\orcidicon[1]{\href{https://orcid.org/#1}{\mbox{\scalerel*{
\begin{tikzpicture}[yscale=-1,transform shape]
\pic{orcidlogo};
\end{tikzpicture}
}{|}}}}

\usepackage{url}

\ifblind
\hypersetup{
  pdfauthor={Anonymous},
  pdftitle={Anonymous Submission},
  pdfsubject={},
  pdfkeywords={}
}
\fi

\ifblind
\author{
	\phantom{\textbf{Bastian Lampe}\textsuperscript{\protect\orcidicon{0000-0002-4414-6947}}, \textbf{Lutz Eckstein}}%
	\thanks{%
		\phantom{\strut\textbf{Institute for Automotive Engineering~(ika),}}\newline
		\phantom{\strut\textbf{RWTH Aachen University, 52074 Aachen, Germany.}}\newline
		\phantom{\strut{\tt\small \textbf{bastian.lampe@ika.rwth-aachen.de,}}}\newline
		\phantom{\strut{\tt\small \textbf{lutz.eckstein@ika.rwth-aachen.de}}}%
	}%
}
\else
\author{
	\textbf{Bastian Lampe}\textsuperscript{\protect\orcidicon{0000-0002-4414-6947}}, \textbf{Lutz Eckstein}%
	\thanks{%
		\strut\textbf{Institute for Automotive Engineering~(ika),\newline RWTH Aachen University, 52074 Aachen, Germany.}\newline
		\strut{\tt\small \textbf{bastian.lampe@ika.rwth-aachen.de, lutz.eckstein@ika.rwth-aachen.de}}%
	}%
}
\fi

\begin{document}
	
\maketitle
\thispagestyle{empty}
\pagestyle{empty}

\begin{abstract}
	The continual assurance of safety and performance of automated driving systems~(ADSs) poses significant challenges. ADSs operate in complex, dynamic, open-world environments allowing a wide range of scenarios, including ones that are rare or not foreseen during initial development. While the incorporation of artificial intelligence~(AI) and machine learning~(ML) technology allows ADSs to learn from data gathered during operation and thus enables them to adapt over time, these approaches come with their own challenges.
	A key advantage of ADSs compared to human drivers is their greater ability to gather data collectively across a fleet of vehicles, or even across multiple fleets operated by different entities, and to learn from this data collectively. Vehicles can share and combine their data to identify additional learning opportunities otherwise missed by individual vehicles.
	This creates new opportunities to tackle the challenges of continual assurance of safety and performance, but requires the implementation of architectures that leverage the collective learning potential. Based on established MLOps principles and existing work in the field of connected automated driving, this paper presents a five-layer architecture for collective learning-enabled MLOps processes for ADSs. The goal of this architecture is to provide a conceptual blueprint for the design and implementation of MLOps processes by fleet operators and other relevant stakeholders. The paper describes the main responsibilities of each layer, their interactions, and how multi-level self-assessments enabled by the architecture can support the detection and reduction of edge cases including black swan~events.
\end{abstract}

\section{Introduction}\label{sec:introduction}

The machine learning operations~(MLOps) paradigm extends the \mbox{DevOps}~\cite{iso.32675-2015.devops.2022} paradigm to the specific challenges of ML-based systems~\cite{sculley.mlops.2015,kreuzberger.mlops.2023}.
When such systems are embedded in automated driving systems~(ADSs), they face additional requirements due to their operation in complex, dynamic, open-world environments, where misuse, failures, or functional insufficiencies can lead to severe harm to people, property, or the environment.

To address these challenges, several regulations and standards have been established. In the EU, the AI~Act~\cite{EC.AIact.2024} provides a uniform legal framework for AI systems, including high-risk AI systems such as ADSs. The AI~Act requires sector-specific legislation to be aligned with its mandatory requirements, such as the Type Approval Framework Regulation (EU)~2018/858~\cite{EC.typeapproval.2018}, the General Safety Regulation~(EU)~2019/2144~\cite{EC.TypeApprovalGeneralSafety.2019}, and implementing rules for ADS type-approval~in~(EU)~2022/1426~\cite{EC.TypeApprovalAutomatedVehicles.2022}. 

Important standards include ISO~26262~\cite{iso.26262.2018} for functional safety~(FuSa), ISO~21448~\cite{iso.21448-2022} for safety of the intended functionality~(SOTIF), and ISO/TS~5083~\cite{iso.5083.2025} for ADS safety design, verification, and validation. 
AI systems are addressed in ISO/PAS~8800~\cite{iso.8800.2024}, which provides a framework for managing AI safety that builds upon ISO~26262 and ISO~21448. UL~4600~\cite{ul4600.2023} describes assessment criteria for safety cases applicable to vehicles operating in autonomous mode and incorporating AI technology.
At the UN level, the NATM Guidelines for Validating ADS~\cite{UN.validationADS.2023} aim to provide a repeatable, objective and evidence-based multi-pillar approach to validating ADS safety. A draft UN Global Technical Regulation~(GTR)~\cite{UN.GTR.ADS.2026} builds on this guidance by defining harmonized ADS requirements and corresponding compliance assessments.
These documents provide important guidelines, however, they do not specify concrete architectures that are able to meet the requirements. While this is intentional to allow for flexibility in implementation, it also means that the question of how to design and implement MLOps processes for ADSs remains open.

\begin{figure}[!t]
	\centering
	\includegraphics[trim=15.5 25 15 5, clip, width=0.815\columnwidth,]{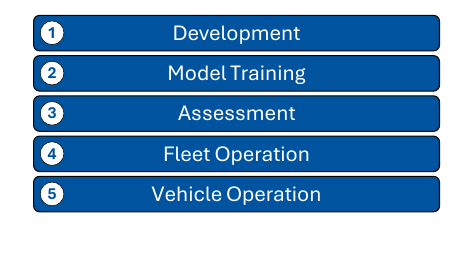}
	\caption{The five layers of the proposed MLOps architecture.}
	\label{fig:layers}
\end{figure}

Both research~\cite{kreuzberger.mlops.2023,johansson.continualLearning.2022,ullrich.vmodel.2024,wagner.safetyCaseFramework.2024,pegasus.vvm.2025,bogdoll.mcityDataEngine.2025} and industry~\cite{webb.waymossafetymethodologiessafety.2020,waymo.safetylifecycle.2023,shalev-shwartz.safetyArchitecture.2024,favaro.determiningabsenceunreasonablerisk.2025} have made important contributions to defining abstractions, requirements or components of MLOps processes that can be applied to ADSs. However, there remains a need for a comprehensive architecture that integrates these contributions into a coherent conceptual blueprint for the implementation of an MLOps process for ADSs.

Moreover, previous work has not yet focused on the potential of vehicle connectivity to expand the detection of learning opportunities beyond edge cases that can be detected in the data of individual vehicles. This paper explains how black swan events as defined by Shalev-Shwartz et al.~\cite{shalev-shwartz.safetyArchitecture.2024} can be detected through collective assessments across multiple vehicles as part of an MLOps process following the presented architecture. 
The contributions of this paper are:
\begin{itemize}
	\item {A five-layer MLOps architecture for ADSs enabling collective, continual learning, and the continual assurance of safety and performance.}
	\item {A description of layer responsibilities and cross-layer interactions that realize the closed MLOps loop.}
	\item {A multi-level self-assessment concept for detecting and reducing edge cases including black swan events.}
\end{itemize}

\section{Related Work}\label{sec:related_work}

\begin{figure}[t]
    \vspace{0.4em}
    \centering
    \includegraphics[width=0.9\columnwidth]{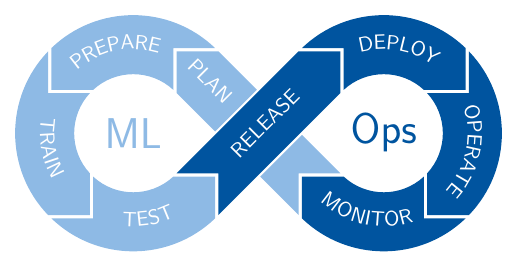}
    \caption{The MLOps lifecycle describes the continuous process of integrating ML model development and model operations to automate and streamline the software delivery pipeline of ML models.}
    \label{fig:mlopsLifecycle}
\end{figure}

The MLOps paradigm extends the DevOps paradigm to the specific challenges of developing and deploying ML models, i.e., for the operationalization of ML products. The MLOps paradigm includes ``\textit{aspects like best practices, sets of concepts, as well as a development culture when it comes to the end-to-end conceptualization, implementation, monitoring, deployment, and scalability of machine learning products}''~\cite{kreuzberger.mlops.2023}. It needs to address ML-specific challenges described by, e.g., Sculley et al.~\cite{sculley.mlops.2015}. Table~\ref{tab:mlop_challenges} summarizes a selection of these key ML-specific challenges.

\begin{table}[h!] 
	\centering
    \caption{Key ML-specific challenges that need to be addressed by the MLOps paradigm.}
    \label{tab:mlop_challenges}
    \begin{tabular}{p{0.18\columnwidth} p{0.71\columnwidth}}
    \toprule
    \textbf{Challenge}    & \textbf{Description}                                                     \\ \midrule
    \raggedright\textbf{Entanglement} & ML models often entangle various input signals and depend on interdependent hyperparameters, resulting in the so-called CACE~principle: Changing Anything Changes Everything. A small change in any dependency can influence all others. \\ \addlinespace[0.3em]
    \raggedright\textbf{Data Drift} & The performance of ML models is affected by data inputs that may change over time, as the external world constantly evolves. Even changes in the input signals that seem beneficial can have negative effects on the overall system performance. \\ \addlinespace[0.3em]
    \raggedright\textbf{Feedback Loops} & Deployed models may directly or indirectly affect each other, and their own future training data, making it difficult to predict their behavior before deployment. \\ \addlinespace[0.3em]
    \bottomrule
    \end{tabular}
\end{table}

The MLOps lifecycle depicted in Fig.~\ref{fig:mlopsLifecycle} is an abstract illustration of a continuous process that integrates ML model development and operations to automate and streamline the delivery of ML models into production environments. The infinity loop represents the continuous cycle that includes a set of iterative phases described in Table~\ref{tab:MLOps_Lifecycle_Overview}.

\begin{table}[t]
	\vspace{0.6em}
    \caption{Description of the key phases within the MLOps lifecycle, divided into ML and Ops sections.}
    \label{tab:MLOps_Lifecycle_Overview}
    \begin{tabular}{p{0.08\columnwidth} p{0.09\columnwidth} p{0.66\columnwidth}}
    \toprule
    \textbf{Section}    & \textbf{Phase}  & \textbf{Description}                                                     \\ \midrule
    \textbf{ML} & \textbf{Plan}   & Analysis of available data, formulation of requirements, aligning development goals with business objectives, and defining relevant KPIs. \\ \addlinespace[0.1em]
                        & \textbf{Prepare}   & Collecting, cleaning, and labeling data, including handling of missing or incorrect data, and ensuring data quality for training. \\ \addlinespace[0.1em]
                        & \textbf{Train}   & Using prepared data to train ML models including model selection, hyperparameter tuning, and experiment tracking. \\ \addlinespace[0.1em]
                        & \textbf{Test}   & Evaluating the performance of trained models on data not used for training using predefined metrics. \\ \midrule
    \textbf{Ops} & \textbf{Release} & Preparing ML models for deployment, including versioning, approvals, auditing, and packaging into artifact repositories. \\ \addlinespace[0.1em]
                        & \textbf{Deploy} & Deploying packaged ML models into production environments. \\ \addlinespace[0.1em]
                        & \textbf{Operate} & Managing the live system, ensuring performance, scalability, uptime, and infrastructure across Dev, QA, and production environments. \\ \addlinespace[0.1em]
                        & \textbf{Monitor} & Continuously tracking model performance, identifying model drift, faults, and anomalies. \\
    \bottomrule
    \end{tabular}
\end{table}

\begin{figure}[b]
    \centering
    \includegraphics[trim=24 329 17 78, clip, width=0.9\columnwidth]{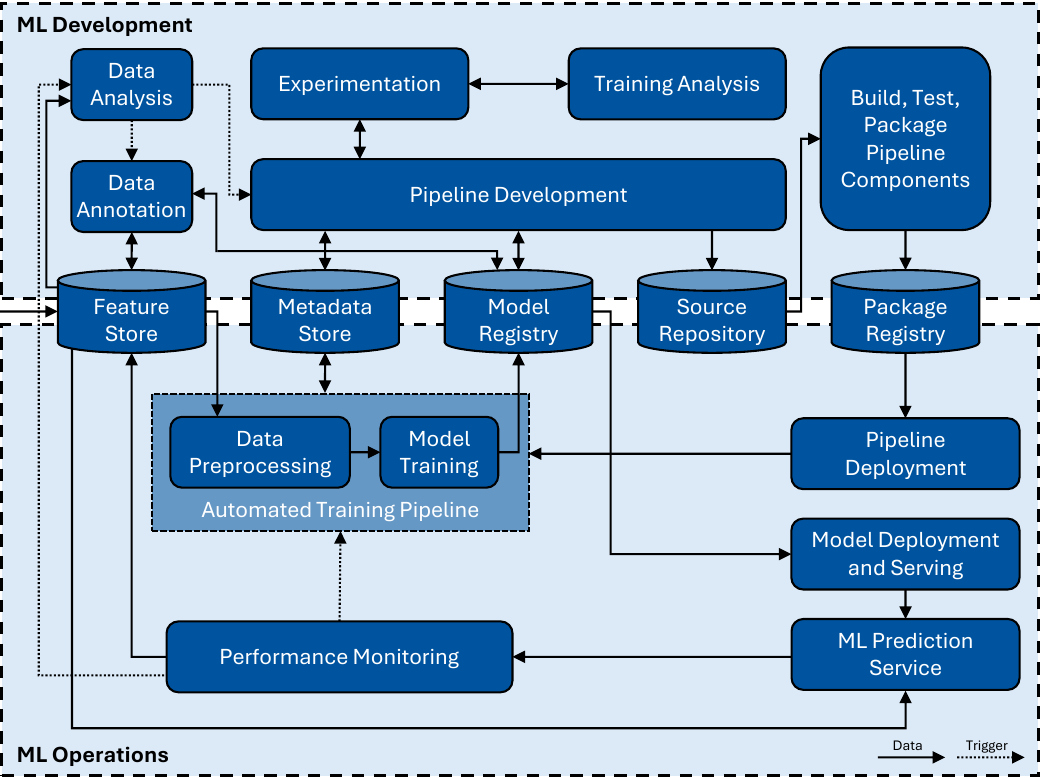}
    \caption{Architecture of a typical MLOps process that realizes the MLOps lifecycle. The figure depicts the most important tasks and storage components, and shows the general data flow between them. The figure builds upon those depicted in~\cite{kreuzberger.mlops.2023, google.MLOps.2024, mlops.MLOpsPrinciples.2024, griciunas.swirlai.2025}.}
    \label{fig:mlops_architecture_sota}
    \vspace{-0.15em}
\end{figure}

Fig.~\ref{fig:mlops_architecture_sota} provides an architecture of a typical MLOps process that realizes the MLOps lifecycle depicted in Fig.~\ref{fig:mlopsLifecycle}. ML models are served and provide a prediction service to consumers of said service. 
The prediction service receives its input data from a feature store. Alternatively, the service may also consume input data directly from the service consumer.

The performance of the prediction service is monitored and if the performance degrades, e.g., due to data drift~(cf.~Table~\ref{tab:mlop_challenges}), an automated training pipeline for retraining the model can be triggered. Model training in this case includes updating model weights to increase model performance, applying the model to validation data to select a model for testing, and testing the model on dedicated testing data to give an unbiased estimate of the model performance. Training runs are tracked in a metadata store and corresponding models are saved to a model registry.

Severe performance degradations or other anomalies in the performance monitoring can also trigger data recordings that can be fed into the feature store, which includes preprocessing steps that transform the raw data into useful features. 

They may also trigger a more thorough data analysis by developers. These may perform exploratory data analysis, adjust the automated training pipeline, request more or different annotated data for training, experiment with new model architectures, or adjust training hyperparameters.\pagebreak

Both the resulting models and the training pipeline itself are analyzed with regard to their performance. Entanglement~(cf.~Tab.~\ref{tab:mlop_challenges}) makes these steps particularly difficult.

Experiments are tracked in a metadata store and corresponding models are saved in a model registry. If the training pipeline performance and the model performance meet their respective requirements, the source code repository of the automated training pipeline is updated and automatically built, tested and packaged. Resulting artifacts are deployed, possibly going through a staging and quality assurance~(QA) environment before. 

Once an improved model generated by the automated training pipeline is pushed to the model registry, the prediction service can be updated with the new model, and the cycle starts anew. 
A key characteristic of this architecture is that developers do not directly produce models to be deployed but rather develop and maintain the automated training pipeline that produces the models and evaluates their~performance.

\section{Challenges specific to ML-based ADSs}

Tab.~\ref{tab:ads_properties} lists challenges of operating ADS-equipped vehicles that require the design of ML systems and of MLOps architectures for ADSs to differ from many conventional, more centralized ML systems and MLOps architectures. 
The combination of these properties makes the design of MLOps architectures for ADSs particularly challenging compared to many other ML-based systems.

\begin{table}[t!]
    \vspace{0.7em}
    \centering
    \caption{Properties of ADS-equipped vehicles that require the design of an MLOps architecture for ADSs to differ from many conventional MLOps processes.}
    \label{tab:ads_properties}
    \begin{tabular}{p{0.18\columnwidth} p{0.71\columnwidth}}
    \toprule
    \textbf{Challenge} & \textbf{Description} \\ \midrule
    \raggedright\textbf{Distributed} & There is not a single or few data center-based endpoints that provide a scalable prediction service to consumers; instead, each ADS-equipped vehicle comes with its own local prediction services, e.g., for object detection, behavior prediction, and behavior planning. \\ \addlinespace[0.1em]
    \raggedright\textbf{Isolated} & There are relatively strict limitations regarding the transmission of data from the prediction service in ADS-equipped vehicles to offboard systems. Sensor data, for example, can only be shared to a very limited degree with current technology. \\ \addlinespace[0.1em]
    \raggedright\textbf{Self-Sufficient} & ADS-equipped vehicles must be able to~--~at least temporarily~--~operate without connectivity to offboard systems, as they need to maintain safety even in the event of a loss of connectivity to these systems. \\ \addlinespace[0.1em]
    \raggedright\textbf{High-Risk} & In the context of the AI~Act, ADS-equipped vehicles can be considered high-risk systems with the potential to cause significant harm and must therefore conform to strict regulatory requirements and standards. \\  \addlinespace[0.1em]
    \raggedright\textbf{Heterogeneous} & There may be different hardware and software configurations in different ADS-equipped vehicles, leading to variations in their capabilities and performance. \\ \addlinespace[0.1em]
    \raggedright\textbf{Limited} & With respect to resources such as computational power, memory, and energy supply, ADS-equipped vehicles are much more limited compared to data centers. \\ \addlinespace[0.1em]
    \raggedright\textbf{Interacting} & ADS-equipped vehicles directly influence each other's operational context, making vehicle functions interdependent in multi-vehicle scenarios. \\
    \bottomrule
    \end{tabular}
\end{table}

\section{Self-Assessment and Learning Opportunities}

The high-risk nature of ADS-equipped vehicles places a strong emphasis on the performance monitoring component of the MLOps process. In the context of MLOps for ADSs, a multi-level performance self-assessment (cf.~Table~\ref{tab:confidence_methods_challenges}) is important for the detection of \textit{learning opportunities}, e.g., when the performance of an ML model is insufficient, and data recordings shall be triggered to allow developers to incorporate gained insights and new data into the continual learning process of affected ML models. 
Self-assessments are also important for maintaining the operational safety of a vehicle by, e.g., triggering a minimal risk maneuver in response to certain operating conditions~\cite{ackermann.safeHalt.2023}. 

\begin{table}[b!]
    \vspace{0.4em}
    \centering
    \caption{Assessment levels in fleets of ADS-equipped vehicles and corresponding challenges.}
    \label{tab:confidence_methods_challenges}
    \begin{tabular}{p{0.18\columnwidth} p{0.71\columnwidth}}
    \toprule
    \textbf{Assessment} & \textbf{Example Challenges} \\ \midrule
    \raggedright\textbf{Service-Level Assessment} &
    \textbf{Limited context}: Can leverage data from an individual service, but cannot account for information beyond the individual task of the service. \newline \textit{Example: Object detector may not have access to information about the road layout.} \\ \addlinespace[0.1em]
    & \textbf{Confidence calibration}: Inability to estimate confidence may correlate with inability to successfully perform the task. \newline \textit{Example: Object detector may be overconfident when encountering a rare object for which the confidence computation has not yet been calibrated.} \\ \addlinespace[0.1em]

    \raggedright\textbf{Subsystem-Level Assessment} &
    \textbf{Correlations of functional insufficiencies}: Functional insufficiencies in multiple redundant components of a vehicle subsystem may be correlated, making it difficult to identify disagreements. \newline \textit{Example: A partially occluded object may be missed by all sensors of the perception subsystem, because they are all affected by a performance insufficiency.} \\ \addlinespace[0.1em]

    \raggedright\textbf{ADS-Level Assessment} &
    \textbf{Error detection delay}: Errors may only become apparent after some time, potentially too late for effective or comfortable risk mitigation. \newline \textit{Example: A false negative detection is only recognized once the emergency braking system~--~which is working with a more rudimentary perception approach and near-field sensors~--~is activated.} \\ \addlinespace[0.1em]

    \raggedright\textbf{Collective-Level Assessment} &
    \textbf{Data volume}: Live transmission of large amounts of data from multiple vehicles may be infeasible. \newline \textit{Example: Sensor data like point clouds may be too large to be transmitted in real-time, making it infeasible to compare the lidar sensor output of multiple vehicles.} \\ \addlinespace[0.1em]
    & \textbf{Correlations of functional insufficiencies}: Functional insufficiencies in multiple vehicles may still be correlated, making it difficult to identify mismatches. \newline \textit{Example: Multiple vehicles may not detect an object that was not present in the training data, i.e., an insufficiency in specification.} \\
    \bottomrule
    \end{tabular}
\end{table}

The architecture of ADSs allows self-assessments to be conducted at different levels. An ADS following a microservices architecture may be composed of different services~(e.g.,~object detection) that are part of different subsystems~(e.g.,~perception) that in turn are part of the overall ADS. Multiple ADS-equipped vehicles form a fleet that may be supported by additional offboard systems. Connectivity between these fleet-level entities also allows collective assessments. Data can be shared across vehicles and offboard systems to be analyzed collectively, e.g., in a cloud backend.\pagebreak

Table~\ref{tab:confidence_methods_challenges} lists different levels at which the assessment of whether data shall be recorded can be performed, and exemplary challenges that may arise at these levels.

To better understand the learning opportunities that arise in the context of a fleet of ADS-equipped vehicles, a \textit{confidence-performance matrix} is introduced in Fig.~\ref{fig:risk_matrix}. It describes different combinations of the level of confidence a system can have in meeting its current functional requirements and whether the requirements are actually met. A ``\textit{system}'' can in this case refer to different levels, as described in Tab.~\ref{tab:confidence_methods_challenges}. The matrix is inspired by the visualization of scenario categories in Fig.~6 of ISO~21448~(SOTIF)~\cite{iso.21448-2022}. There, scenarios are categorized based on whether they are \textit{known} or \textit{unknown}, and whether they are \textit{hazardous} or \textit{not hazardous}. In ISO~21448, the knowledge dimension reflects the state of knowledge within the SOTIF lifecycle. In contrast, the confidence-performance matrix in Fig.~\ref{fig:risk_matrix} refers to the \textit{runtime state} of the system itself.

Reliable system behavior is characterized by a high performance and high confidence. There is no immediate need to improve the system in this case. Data recordings may be triggered, but are not required. A high performance coupled with a low confidence will lead to defensive behavior. Since the system does not know whether its actual performance is sufficient, it shall employ safety functions. Data recordings can be triggered in this case, since the low confidence can be used as a trigger, and they should also be triggered, because defensive behavior may negatively affect user satisfaction, but can also have further negative consequences.

A low performance combined with low confidence may result in a violation of safety requirements. This combination represents \textit{hazardous behavior} and can correspond to a \textit{hazardous event}, i.e., an ``\textit{event that may result in harm}''~\cite{iec.61508.2010}; therefore, safety functions shall be employed. If such an event occurs in a scenario that can lead to harm and the system is unable to control the hazardous event, such behavior may result in actual harm. Data recordings again can and should be triggered in this case. 

Conversely, low performance with high confidence constitutes high-risk behavior, as no safety functions are activated to achieve a safe state. In this case, the system is unaware of its own limitations and may cause harm. The likelihood of harm can be greater compared to hazardous behavior, since safety functions are not engaged, increasing the probability that hazardous events will not be controlled, even if they could be controlled if the system was aware of them.

High-risk behavior is associated with an \textit{unknown unknown} or \textit{black swan} event when viewed on an individual assessment level. \textit{Black swan} events refer to rare events in which a system fails while remaining unaware of the failure, and where the failure is not reproducible or reproducibility is unknown~\cite{shalev-shwartz.safetyArchitecture.2024}. Although rare at vehicle level, such events can~--~like regular edge cases~--~still be numerous at fleet level. An unknown unknown on one assessment level may be transformed into a \textit{known unknown} by moving the assessment to a higher level, e.g., by aggregating data and confidence measures from multiple redundant services~(subsystem-level assessment), or by aggregating data and confidence measures from multiple vehicles~(collective-level assessment). However, there is no guarantee this is always possible.

\colorlet{reliable}{greenLightRWTH}
\colorlet{risk}{redLightRWTH}
\colorlet{defensive}{ikayellow}
\colorlet{hazard}{orangeLightRWTH}

\newcommand{\opbox}[2]{%
    \parbox[c][3.0cm][c]{3.0cm}{\centering\Large \textbf{#1}\\ \small\textit{#2}}%
}
\newcommand{\texts}{\opbox{Reliable \\ Behavior}{Sufficient performance with adequate confidence.}}
\newcommand{\textw}{\opbox{High-Risk \\ Behavior}{System does not react to hazards, leading to increased risk.}}
\newcommand{\texto}{\opbox{Defensive \\ Behavior}{System employs safety functions, maintaining safety.}}
\newcommand{\textt}{\opbox{Hazardous \\ Behavior}{System employs safety functions, minimizing risk.}}

\begin{figure}[t!]
    \vspace{0.5em}
  \centering
  \resizebox{0.9\columnwidth}{!}{%
  \begin{tikzpicture}[>=latex, line cap=round, rounded corners=0.5pt, x=8.5cm, y=8.5cm]
    \def\sx{0.50}
    \def\sy{0.50}

    \draw[very thick, white] (0,0) rectangle (1,1);

    \fill[reliable!50]  (0,\sy) rectangle (\sx,1);   
    \fill[risk!50]      (\sx,\sy) rectangle (1,1);   
    \fill[defensive!50] (0,0)   rectangle (\sx,\sy); 
    \fill[hazard!50]    (\sx,0) rectangle (1,\sy);   

    \fill[blueLighterRWTH, sharp corners] (0.002,1) rectangle (\sx,1.08);
    \fill[blueLighterRWTH, sharp corners] (\sx,0.999) rectangle (0.998,1.08);
    \node[font=\large, text=black] at (0.25,1.03) {High Performance};
    \node[font=\large, text=black] at (0.75,1.035) {Low Performance};

    \fill[blueLighterRWTH, sharp corners] (-0.08,\sy) rectangle (0,0.998);
    \fill[blueLighterRWTH, sharp corners] (-0.08,0.002) rectangle (0,\sy);
    \node[font=\large, rotate=90, text=black] at (-0.035,0.75) {High Confidence};
    \node[font=\large, rotate=90, text=black] at (-0.04,0.25) {Low Confidence};

    \draw[line width=0.5pt, white] (-0.08,\sy) -- (-0.000,\sy);
    \draw[line width=0.5pt, white] (\sx,1.08) -- (\sx,1.000);
    \draw[line width=0.5pt, black] (\sx,0.001) -- (\sx,0.999);
    \draw[line width=0.5pt, black] (0.001,\sy) -- (0.999,\sy);
    \draw[line width=0.5pt, black] (0.000,0.001) -- (0.999,0.001);
    \draw[line width=0.5pt, black] (0.001,0.999) -- (0.999,0.999);
    \draw[line width=0.5pt, black] (0.001,0.999) -- (0,0.001);
    \draw[line width=0.5pt, black] (0.999,0.999) -- (0.999,0.001);

    \node at (0.25, 0.75) {\texts}; 
    \node at (0.75, 0.75) {\textw}; 
    \node at (0.25, 0.25) {\texto}; 
    \node at (0.75, 0.25) {\textt}; 
  \end{tikzpicture}%
  }
  \caption{The confidence-performance matrix describes different combinations of the level of confidence a system has in meeting its current functional requirements and the degree to which the requirements are actually met. The operation of the system can be classified into four categories: reliable behavior, defensive behavior, hazardous behavior, high-risk behavior. The colors indicate the associated level of risk: green~(low risk) to red~(high-risk). High-risk behavior corresponds to particularly important learning opportunities for the continual improvement of ADS performance.}
  \label{fig:risk_matrix}
  \vspace{1.0em}
    \centering
    \resizebox{0.9\columnwidth}{!}{%
    \begin{tikzpicture}[>=latex, line cap=round, rounded corners=0.5pt]

    \begin{scope}[shift={(-2.7,0)}, x=5.4cm, y=5.4cm]
        \draw[line width=0.6pt, black] (0,0) rectangle (1,1);
        \fill[white] (0.03,0.03) rectangle (0.97,0.97);
        \filldraw[fill=reliable!50, draw=black, line width=0.4pt] (0.05,0.05) rectangle (0.64,0.95);
        \filldraw[fill=defensive!50,draw=black, line width=0.4pt] (0.68,0.05) rectangle (0.95,0.45);
        \filldraw[fill=hazard!50,   draw=black, line width=0.4pt] (0.68,0.49) rectangle (0.95,0.77);
        \filldraw[fill=risk!50,     draw=black, line width=0.4pt] (0.68,0.81) rectangle (0.95,0.95);
    \end{scope}

    \begin{scope}[shift={(7.3,0)}, x=5.4cm, y=5.4cm]
        \draw[line width=0.6pt, black] (0,0) rectangle (1,1);
        \fill[white] (0.03,0.03) rectangle (0.97,0.97);
        \filldraw[fill=reliable!50, draw=black, line width=0.4pt] (0.05,0.05) rectangle (0.72,0.95);
        \filldraw[fill=defensive!50,draw=black, line width=0.4pt] (0.76,0.05) rectangle (0.95,0.45);
        \filldraw[fill=hazard!50,   draw=black, line width=0.4pt] (0.76,0.49) rectangle (0.95,0.77);
        \filldraw[fill=risk!50,     draw=black, line width=0.4pt] (0.76,0.81) rectangle (0.95,0.95);
    \end{scope}

    \node[font=\bfseries\large, align=center] at (0.0,-0.6) {Current\\Software Release};
    \node[font=\bfseries\large, align=center] at (10.0,-0.6) {Goal for Next\\Software Release};

    \draw[very thick,->] (3.1,2.4) -- (7.0,2.4)
        node[midway, above=0.5pt, align=center, font=\bfseries\large] {Collective,\\Continual Learning};

    \end{tikzpicture}%
    }
    \caption{The relative shares of the four behavior classes change through phases of collective, continual learning. The total colored area corresponds to the entire operational domain. The intended result is a larger share of reliable behavior~(green) and smaller shares of defensive behavior~(yellow), hazardous behavior~(orange), and high-risk behavior~(red).}
    \label{fig:risk_matrix_evolution}
\end{figure}

Fig.~\ref{fig:risk_matrix} introduces \textit{confidence} as a second dimension next to the operational system performance because, especially in high-risk systems, functional insufficiencies should be detected \textit{before} they lead to harm, e.g., when triggering conditions leading to hazardous behavior do not lead to harm, because the scenario lacks the necessary~conditions. 

If an ADS-equipped vehicle enters an intersection against a red light, but there is no crossing traffic present, a data recording should still be triggered to allow the continual learning process to incorporate data of this event such that the system shows correct behavior in the future when the same or a similar scenario occurs. Preferably, a data recording is triggered even when the local perception stack is highly confident in an incorrect traffic-light interpretation. However, if the ADS uses only its own local (un-)certainty to determine whether a data recording should be triggered, this event may be missed; hence collective assessments are required. Such collective assessments can combine observations from multiple vehicles and offboard systems to identify inconsistencies that remain invisible at the level of a single vehicle.

In reality, the class boundaries depicted in Fig.~\ref{fig:risk_matrix} are not necessarily discrete, but continuous. Nonetheless, the matrix provides a useful abstraction to describe learning opportunities that arise in fleets of ADS-equipped vehicles. The figure also illustrates the dilemma between trying to minimize the aggregation of data on higher levels due to, e.g., resource limitations or privacy concerns, and the maximization of detected learning opportunities. 

Even if confidence levels in an ADS are high, there may still be a need to aggregate data and confidence measures on a higher assessment level. The need may even be highest in these cases, because the system may not be aware of its own limitations and would not employ safety functions that aim to achieve a safe state. Hence, though seemingly unintuitive, the most important learning opportunities for collective, continual learning with collective assessments may arise in scenarios where an individual ADS shows a high confidence in its current performance.

Fig.~\ref{fig:risk_matrix_evolution} illustrates a central intended result of collective, continual learning within an MLOps process, i.e., to increase the ratio of reliable behavior in the operational domain~(OD) of the ADS, while decreasing the ratio of high-risk, hazardous, and defensive behavior. This may occur not only by increasing the performance of ML-based ADSs, but also through better confidence calibration.

\section{Five-Layer MLOps Architecture}

\begin{figure}
    \vspace{0.3em}
    \centering    
    \includegraphics[trim=23 40 15 3, clip, width=0.95\columnwidth]{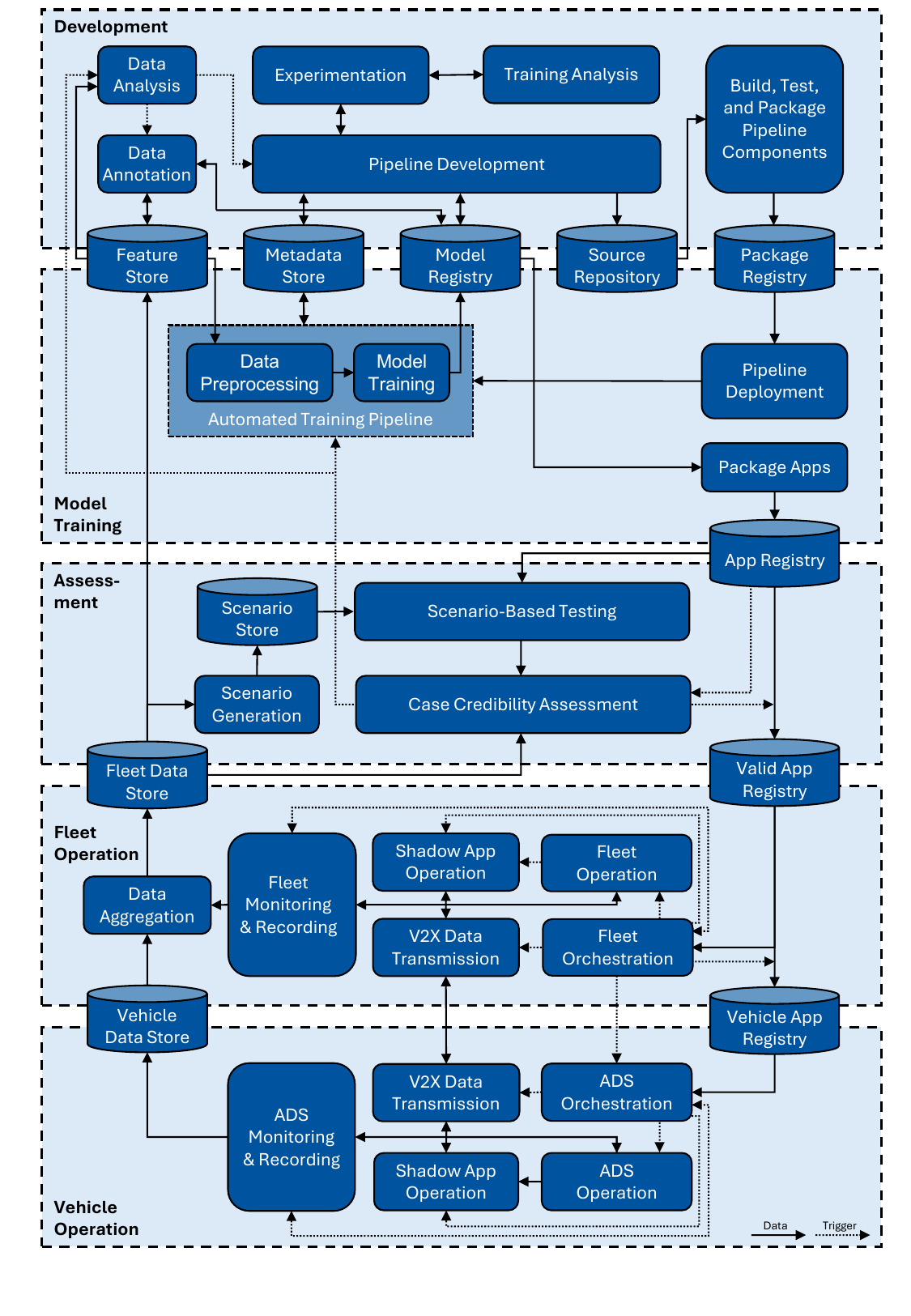}
    \caption{Five-layer architecture of an MLOps process for fleets of ADS-equipped vehicles enabling collective, continual learning. The figure depicts the most important tasks and central storage components.}
    \label{fig:mlops_architecture_ads}
\end{figure}

In contrast to a standard MLOps architecture~(cf.~Fig.~\ref{fig:mlops_architecture_sota}), the MLOps architecture for ADSs is structured into five layers rather than the conventional two-layer split between ML development and ML operations. Note that large parts of the functional architecture depicted in Fig.~\ref{fig:mlops_architecture_ads} would be the same in a general DevOps architecture for ADSs that includes both ML-based and non-ML-based applications.

\subsection{Development Layer}\label{par:ml_development_layer}
The primary purposes of the development layer are the annotation of data for building training, validation, and testing datasets, and the development of training pipeline components for execution in the model training layer. 

Specifically, this layer (1)~adds and refines data annotations, potentially leveraging models from the model registry for active learning or annotation assistance; (2)~performs detailed data analysis, e.g., when triggered by the assessment layer; (3)~conducts experimentation; (4)~analyzes training runs on development datasets, primarily to develop components for the automated training pipeline, with source code maintained in corresponding source repositories; and (5)~builds, tests, and packages training pipeline components for storage in the package registry. 

Similar to a typical MLOps architecture, this layer does not directly generate models for deployment, but rather focuses on the development of training pipeline components and dataset development.

\subsection{Model Training Layer}\label{par:ml_training_layer}
In contrast to the standard MLOps pipeline depicted in~Fig~\ref{fig:mlops_architecture_sota}, this layer is only dedicated to the automated generation and continual improvement of applications that incorporate ML models. It does not operate prediction services; those are handled by the operations layers. 

Concretely, the model training layer (1)~retrieves data samples from the feature store; (2)~performs data preprocessing; (3)~trains and evaluates models~--~optionally initialized from pretrained models retrieved from the model registry~--~within an automated training pipeline that is (4)~automatically deployed from packaged training pipeline components in the package registry.
It (5)~records training metadata and statistics in the metadata store; (6)~registers trained models (with provenance and metrics) in the model registry; and (7)~packages models into applications with explicit description of inputs and outputs, capabilities, and performance profiles. 

Packaged applications are stored in the application registry and are gated by the assessment layer. The majority of aspects in the model training layer remain similar to the standard MLOps pipeline.
\pagebreak
\subsection{Assessment Layer}\label{par:safety_assessment_layer}
The impact of trained models on the performance of an ADS cannot be sufficiently evaluated by the model training layer alone, because ML-based applications are part of a complex system of systems producing emergent behavior that cannot be sufficiently predicted by measuring the performance of individual models. Therefore, evaluation shall also be performed in the context of the entire system of systems or a suitable model thereof. This layer primarily gates the deployment of applications to the operations layers by assessing whether the applications remain, e.g., acceptably~safe.

For this purpose, this layer accesses data from the fleet data store to (1)~generate scenarios for the purpose of (2)~conducting scenario-based testing and to (3)~continually analyze whether, e.g., a safety case for the ADS is still credible, i.e., the safety case still sufficiently argues for the absence of unreasonable risk. 

Analogous to this \textit{Safety} Case Credibility Assessment~(CCA) described in~Schnelle~et al.~\cite{schnelle.assessingSafetyCase.2025}, there should be a \textit{User Acceptance} CCA to maintain and improve user satisfaction over the ADS lifetime. The same applies, e.g., to \textit{security}, \textit{privacy}, \textit{ethical behavior}, and \textit{environmental impact}. Each domain-specific CCA should at least assess the credibility of the domain-specific argument, and of the domain-specific evidence, as proposed by~Schnelle~et~al.~\cite{schnelle.assessingSafetyCase.2025}.

A Safety CCA shall run periodically and/or be triggered by events, e.g., when new or updated applications are available in the application registry, when measured or predicted Safety Performance Indicators~(SPIs) are updated and the two no longer sufficiently align, or when significant edge cases or anomalies were detected and recorded in the fleet data store. Additional triggers may include updated regulations or standards, changes to the operational design domain~(ODD), changes to the vehicle hardware, external reports of safety incidents that were not captured in the fleet data store, changes to the OD, changes to external technical systems on which the ADS relies, changes to the tools used in a CCA, and societal changes relating to the question of what is considered acceptably safe. A CCA is not limited to the results of scenario-based testing. In the context of ML systems, it shall also evaluate possible dataset insufficiencies~(cf.~Jeyachandran~et~al.~\cite{jeyachandran.oasiss.2025} and~ISO/PAS~8800~\cite{iso.8800.2024}). 

Criteria to determine the absence of unreasonable risk are presented by, e.g., Favaro et al.~\cite{favaro.absenceOfUnreasonableRisk.2025}. When scenario-based testing is employed, tests may range from tests of individual services to tests involving the entire system of systems. Open-loop tests including replay/reprocessing tests and replay-to-sim tests; semi-closed-loop tests such as adaptive replay-to-sim tests~\cite{weber.adaptiveReplayToSim.2025} and advanced replay-to-sim tests~\cite{schuldes.deliverable13VVM.2023}; and closed-loop tests can be used. Additionally, tests may include hardware-in-the-loop, software-in-the-loop, and human-in-the-loop testing.

Scenarios used for testing shall reflect already encountered real-world scenarios that can be continually extracted from the fleet data store, relevant variations of these scenarios, and other relevant scenarios that have not yet been encountered and recorded, e.g., those sourced from expert knowledge. The fleet data store may also contain recordings of data produced by so-called shadow mode applications, which are applications that can be tested in parallel with the existing applications, but their output is not used in downstream services that are part of the causal chain affecting vehicle behavior. Additionally, the fleet data store may also contain data from which safety-related metadata can be extracted, e.g., leading SPIs such as frequency of near-misses and frequency of emergency braking, or lagging SPIs such as the frequency of collisions~\cite{ul4600.2023}.

The Safety CCA of this layer may (4)~trigger the automated training pipeline if, e.g., a retraining with new data is expected to improve the safety performance associated with the ADS. It may also (5)~trigger a more detailed analysis and subsequent actions in the development layer if, e.g., the safety case is violated, for instance because the predicted item performance substantially deviates from the actual performance as measured by SPIs. Additionally, it may (6)~trigger the transfer of applications from the application registry to the validated application registry if the safety case is still valid for new or updated applications as part of the overall system of systems.

The Safety CCA may also (7)~revoke the validated status of applications or of the complete ADS which were previously determined acceptably safe. Additionally, (8)~applications may be tagged with a label, e.g., \textit{shadow mode} if an application is suitable for open-loop real-world testing, enabling underlying layers to deploy it in shadow mode. Additional labels may specify deployment strategies and operational guardrails (e.g., OD constraints).

Finally, (9)~the Safety CCA itself should be continually updated based on, e.g., the results of the Safety CCA or new regulations and standards. This meta-analysis of the safety case shall include the analysis of assumptions, SPIs, tools and data used to analyze the validity of the safety case. This includes the analysis of the adequacy of the continuous DevOps/MLOps process itself.

For more details on ADS safety cases, refer to, e.g., the results of the VVM~Project~\cite{pegasus.vvm.2025} or to Waymo's approach to building a safety case~\cite{waymo.safetylifecycle.2023}. For more details on the continual safety assessment of ML systems, refer to~\cite{iso.8800.2024, ul4600.2023,wagner.safetyCaseFramework.2024, schleiss.continuousAssurance.2022, burton.mlAssurance.2023, ratiu.safetyPerformanceIndicators.2024}.

\subsection{Fleet Operation Layer}\label{par:fleet_operation_layer}
The fleet operation layer supports the vehicle operation layer by (1)~exchanging live data with the vehicle operation layer. This enables (2)~additional supportive, cooperative, and collective services to aid vehicles in their operation. This may include providing a digital twin service that integrates data from various sources, including vehicles~\cite{vanKempen.digitalTwin.2023,irfan.transportationDigitalTwin.2024}. To support this digital twin and other purposes, possibly connected, sensor-equipped roadside infrastructure may be~used~\cite{cress.intelligentTransportationSystems.2024}. 

Not all tasks of this layer can be, or are allowed to be, fully automated. Control centers with human operators are required to oversee the operation of the fleet~\cite{EC.TypeApprovalAutomatedVehicles.2022} and intervene or assist if necessary. They perform driving-related tasks such as remote assistance, driving, or intervention, as well as non-driving-related tasks such as communication with, e.g., passengers, and emergency~handling~\cite{wolf.teleoperationControlCenter.2025}.

Similar to the vehicle operation layer, the fleet operation layer may (3)~run shadow mode fleet applications to test new or updated software open-loop. The fleet operation layer is also responsible for (4)~orchestrating the operation of different tasks and applications at fleet level~\cite{lampe.robotkube.2023,zanger.applicationmanagement.2025,reiher.eventDetection.2025}, including the orchestration of tasks across multiple vehicles or other entities in the C-ITS. This also involves data recordings triggered on the collective level.

If applications are tagged for special deployment in the assessment layer, the fleet operation layer is responsible for orchestrating corresponding deployment strategies. These strategies may include \textit{A/B testing}~\cite{mattos.ABTesting.2020}, i.e., deploying two versions of an application to different parts of the fleet and comparing their performance, \textit{canary testing}, i.e., deploying a new version of an application to a small subset of the fleet to, e.g., limit the impact of potential issues, \textit{rolling updates}, i.e., incrementally making updates available to a larger subset of the fleet while monitoring for issues before proceeding further, \textit{OD-limited testing}, i.e., allowing fleet and vehicle orchestration layers to deploy applications only in specific operational domains, \textit{controlled deployment testing}, i.e., allowing fleet and vehicle orchestration layers to deploy applications only on specific vehicles used by the fleet operator for, e.g., test track testing or field-testing with a safety driver.

Fleet monitoring~(5)~is another central task that can trigger data recordings and/or fleet orchestration tasks in the event of, e.g., detected anomalies, edge cases, and functional insufficiencies. Further, the fleet operation layer is responsible for (6)~aggregating data recordings of fleet and vehicle operation layers, possibly enriching the data with additional information, curating data, deriving statistics, and storing data in the fleet data store. Finally, it is responsible for (7)~triggering OTA updates that move applications from the validated applications registry to vehicle application registries when new or updated validated applications are available and vehicles are eligible and capable of receiving~them.

\subsection{Vehicle Operation Layer}\label{par:vehicle_operation_layer}
The vehicle operation layer is responsible for operating individual vehicles. It (1)~orchestrates applications (cf.~Kampmann~\cite{kampmann.dynamicServiceOrientedArchitecture.2023}) available in the vehicle application registry as currently required, (2)~runs the applications that form the vehicle-based ADS responsible for the dynamic driving task~(DDT), including possible safety functions implementing, e.g., minimal risk maneuvers~(cf.~Ackermann~\cite{ackermann.safeHalt.2023}), interactions with remote operators, interactions with passengers via internal HMI, and interactions with the environment via external HMI; it (3)~runs shadow mode vehicle applications, (4)~exchanges live data with the fleet operation layer, (5)~monitors vehicle-based data to record relevant events, especially in case of anomalies, edge cases, ODD boundary approaches/exits, detected functional insufficiencies, and activations/re-initializations and deactivations of the ADS; and triggers the orchestrator when a reconfiguration of the ADS is required. Finally, it (6)~stores recordings in the local vehicle data store for later transmission to the fleet data store via the fleet operation layer.

Fig.~\ref{fig:mlops_architecture_ads} depicts the MLOps architecture for a single fleet of vehicles. It assumes that data can be shared between all layers for the purpose of improving and assuring the overall system performance, i.e., the organization operating the MLOps process holds all necessary access rights to the~data. 
In a multi-fleet case, where multiple organizations operate their own fleets but nonetheless share some components of some of the layers, or in a single-fleet case where different layers~--~or parts thereof~--~are operated by different organizations, the data flow would be more restricted. Federated learning approaches could be applied in these cases.

\section{Conclusion}

The presented five-layer architecture of an MLOps process for ADSs provides a conceptual blueprint for fleet operators and other stakeholders to implement such a process, enabling them to continually assure the safety and performance of their fleet of ADS-equipped vehicles. The architecture aims to adapt standard MLOps architectures to the specific requirements and challenges associated with ADSs, and to leverage the possibilities of collective assessments to detect and reduce black swan events, other functional insufficiencies, and failures. 

By enabling collective, continual learning for ML-based ADSs, the MLOps architecture aims to increase the ratio of reliable behavior in the operational domain of an ADS over time, while decreasing the ratio of high-risk, hazardous, and defensive behavior.

The detailed descriptions of each layer's tasks and responsibilities, and of cross-layer interactions, provide a comprehensive starting point for implementing a corresponding MLOps process.
Future work will include such an implementation to empirically evaluate the architecture and to identify potential gaps and areas for further improvement. 

Additionally, a more detailed mapping of architecture components to the extensive requirements of regulations and standards is necessary to evaluate whether the architecture needs to be adapted or extended to meet all requirements. At the same time, further research is required to determine whether current regulations and standards should be extended to better support collective, continual learning for ML-based~ADSs.

Finally, future research should investigate the applicability of the presented architecture to other classes of complex, high-risk physical AI systems.

\section*{Acknowledgements}

\newcommand{\acktext}{This work was funded by the European Union under the Horizon Europe programme in the AIGGREGATE project (Grant Agreement No. 101202457).
Furthermore, this research received funding from the German Federal Ministry of Research, Technology and Space within the "autotech.agil" project~(FKZ~01IS22088A).}

\ifblind
\phantom{\strut This work was funded by the European Union under the Horizon Europe programme.}\par
\phantom{\strut The funding was provided in the AIGGREGATE project (Grant Agreement No. 101202457).}\par
\phantom{\strut Furthermore, this research received funding from the German Federal Ministry}\par
\phantom{\strut of Research, Technology and Space within the "autotech.agil" project}\par
\phantom{\strut (FKZ~01IS22088A).}\par
\phantom{\strut }%
\else
\acktext
\fi

\pagebreak
\bibliographystyle{IEEEtran}
\bibliography{root} 
	
\end{document}